\documentclass[11pt]{article}

\usepackage{acl}

\usepackage{times}
\usepackage{latexsym}
\usepackage{subcaption}

\usepackage[T1]{fontenc}

\usepackage[utf8]{inputenc}

\usepackage{microtype}

\usepackage{inconsolata}

\usepackage{graphicx}
\newcommand{\header}[1]{\vspace{2mm}\noindent\textbf{#1}}

\usepackage[inline]{enumitem}
\usepackage{pbox}
\usepackage{balance}
\newlist{inlinelist}{enumerate*}{1}
\setlist*[inlinelist,1]{label=\roman*),itemjoin={{, }},itemjoin*={{, and }}}
\usepackage{url}
\usepackage{multirow}
\usepackage{multicol}
\usepackage{booktabs}
\usepackage{tikz}
\usepackage{xcolor}
\usepackage{lipsum}
\usepackage{acronym}
\usepackage{subcaption}
\usepackage[inline]{enumitem}
\usepackage{xspace}
\usepackage{arydshln}
\usepackage{array}
\usepackage{adjustbox}
\usepackage{newfloat}
\DeclareFloatingEnvironment[fileext=lop, listname={List of prompts}, 
                            name=Prompt, placement=h]{prompt}
\usepackage{listings}
\usepackage{xcolor}
\lstdefinelanguage{PolicyPrompt}{
  literate=
    {<think>}{{{\color{green!60!black}<think>}}}{7}
    {</think>}{{{\color{green!60!black}</think>}}}{8}
    {<search>}{{{\color{orange!80!black}<search>}}}{8}
    {</search>}{{{\color{orange!80!black}</search>}}}{9}
    {<information>}{{{\color{blue!80!black}<information>}}}{13}
    {</information>}{{{\color{blue!80!black}</information>}}}{14}
    {<answer>}{{{\color{red!70!black}<answer>}}}{8}
    {</answer>}{{{\color{red!70!black}</answer>}}}{9}
    {conversation\_text}{{{\color{red!90!black}conversation\_text}}}{13}
    {last\_user\_utterance}{{{\color{red!90!black}last\_user\_utterance}}}{19}
    {top20\_retrieved\_text}{{{\color{red!90!black}top20\_retrieved\_text}}}{19}
    {top5\_retrieved\_text}{{{\color{red!90!black}top5\_retrieved\_text}}}{19}
}

\usepackage{acl}
\usepackage{times}
\usepackage{latexsym}
\usepackage{amsmath}
\usepackage{amssymb}
\usepackage[T1]{fontenc}
\usepackage[utf8]{inputenc}
\usepackage{microtype}
\usepackage{inconsolata}
\usepackage{graphicx,xcolor}
\usepackage{booktabs}
\usepackage{pifont}

\title{uva-irlab-conv at SemEval-2026 Task 8: Multi-Turn RAG with Learned Sparse Retrieval and Listwise Reranking}

\author{
Simon Lupart \quad
Kidist Amde Mekonnen \\
{\bf Zahra Abbasiantaeb \quad
Mohammad Aliannejadi} \\
University of Amsterdam \\
Amsterdam, The Netherlands \\
\texttt{\{s.c.lupart, k.a.mekonnen, z.abbasiantaeb, m.aliannejadi\}@uva.nl}
}

\begin{document}
\maketitle
\begin{abstract}

This report describes our participation in SemEval-2026 Task 8 on multi-turn retrieval and question answering. The task evaluates conversational systems across four domains (finance, cloud documentation, government, Wikipedia), and includes unanswerable queries where the available collection does not contain sufficient evidence to produce a complete response.
We propose a multi-turn retrieval-augmented generation pipeline that combines learned sparse retrieval with LLM–based reranking and generation. Using sparse retrieval as the primary retrieval method, we leverage its strong generalization across domains. In addition, we make use of the long-context capabilities of LLMs for conversational query rewriting, pointwise and listwise reranking, and generating the final response, each conditioned on the full conversational history.
This multi-step design enables effective integration of conversational context throughout retrieval and generation, improving robustness across domains.

\end{abstract}

\section{Introduction}

SemEval-2026 Shared Task 8 is based on the Multi-Turn RAG (MTRAGEval) benchmark~\cite{Rosenthal2026MTRAGEval,rosenthal2026mtragunbenchmarkopenchallenges}, which evaluates conversational search and question answering (QA) systems across multi-turn interactions. The benchmark focuses on complex information needs arising in dialogue, where conversational context evolves and may introduce topic shifts, ambiguity, and noisy historical information that systems must resolve~\cite{laban2026llms,theoreticalframeCS}. In addition, the task is motivated by the increasing role of LLMs as information access interfaces~\cite{chatterji2025people,dalton2022conversational}. Systems are required not only to retrieve and synthesize evidence across multiple turns, but also to recognize and appropriately handle unanswerable or underspecified queries when insufficient supporting information is available. This includes the ability to signal uncertainty or request clarification when needed.

Our approach follows a multi-stage retrieval-augmented generation (RAG) pipeline combining conversational query rewriting, learned sparse retrieval (LSR), and LLM-based reranking. We first rewrite the latest user utterance into a standalone query using the full conversation history~\cite{elgohary-etal-2019-unpack}. The rewritten query is then used for retrieval with LSR~\cite{lsrzamani,unilsrthong}, which integrates neural semantic representations with lexical sparsity to provide robust cross-domain generalization~\cite{spladev1}. An initial reranking stage selects a set of candidate passages, which are subsequently refined through LLM-based listwise reranking that leverages the full conversational context for finer-grained evidence selection~\cite{sun2023chatgpt}. Finally, the top-ranked passages are used to generate responses in a zero-shot RAG setting~\cite{gao2023retrieval}.

Participation in the shared task provides several insights. Our system demonstrates strong retrieval effectiveness, \textit{ranking 2nd out of 38 teams} with an nDCG@5 score of 0.5475. Performance varies across domains, with the finance domain proving the most challenging and the Wikipedia-based domain the easiest. In contrast, generation performance is lower (\textit{23/26 and 20/29}), primarily because our submission does not explicitly model unanswerable scenarios. Nevertheless, qualitative analysis shows that the system remains faithful when evidence from the collection is insufficient.

\section{Background}

Conversational search and QA have been studied through several shared tasks~\cite{Dalton2020CAsT2T,mohikat23,mohikat24,gohsen2025user} and offline datasets~\cite{anantha-etal-2021-open,adlakha-etal-2022-topiocqa}, but most existing resources focus primarily on general domains (where LLMs can often rely on parametric knowledge) or focus on different aspects of the evaluation (e.g. personalization). In contrast, the MTRAG benchmark emphasizes domain-specific collections (FiQA~\cite{maia201818}, ClapNQ~\cite{rosenthal2025clapnq}, and two novel collections on Cloud documentation and government, all in English). It also explicitly evaluates unanswerable scenarios and response faithfulness, placing stronger requirements on retrieval and grounding~\cite{es-etal-2024-ragas}. 
The organizers defined three sub-tasks. \textbf{Task A} focuses on conversational search. \textbf{Task C} evaluates response generation in a standard RAG setting, where answers are generated from retrieved passages. Finally, \textbf{Task B} serves as an oracle RAG, where the response generation system has access to gold-labeled passages.

Our submission builds on previous participation in shared tasks using LSR in conversational search~\cite{lupart2025uvairlab,lupart2024irlab}. More specifically, we rely on query rewriting, a dominant paradigm in conversational QA, reformulating context-dependent utterances into standalone queries compatible with standard retrieval models~\cite{vakulenko2021question}, further enhanced now with the use of LLMs~\cite{mao2023largelanguagemodelsknowllm4cs,mo2024chiqcontextualhistoryenhancement,lupartdisco,lupart2025varia}. While unified approaches that jointly model retrieval and reasoning have also been proposed~\cite{mo-etal-2025-uniconv,abbasiantaeb2024mq4cs,lupart2025chatr1,mo2026agentic,mekonnen}, they typically require substantial supervised data or task-specific training. We therefore adopt a lightweight zero-shot strategy based on LLM-based query rewriting and retrieval-augmented generation. More advanced RAG pipelines, including query decomposition and self-reflection over retrieved passages~\cite{yao2022react,asai2023self,trivedi2023interleaving}, are left for future work.

\section{System Overview}

\begin{figure}[t!]
    \centering
    \includegraphics[width=\linewidth]{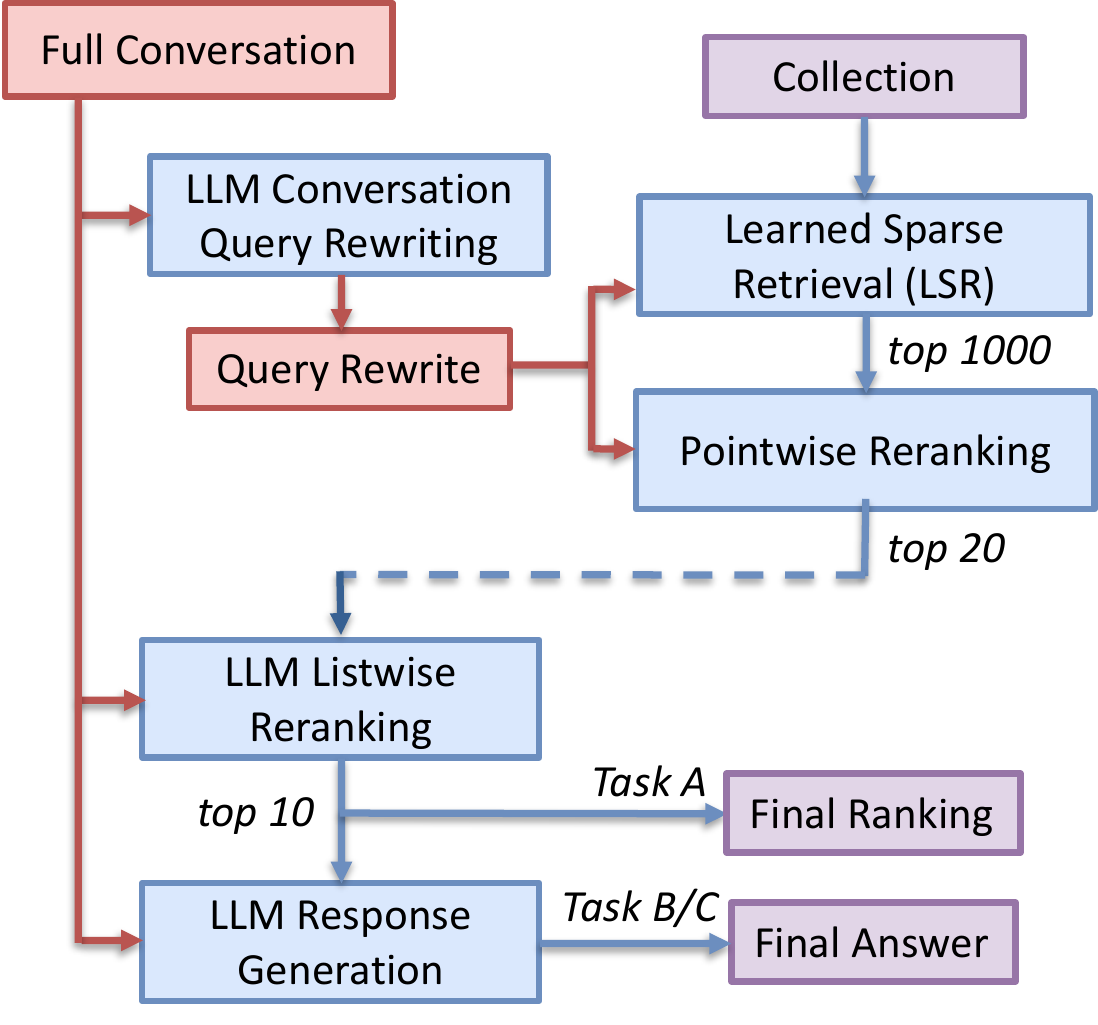}
    \caption{
    Overview of our submission. Early stages rely on query rewriting, while later stages leverage the full conversation history for more fine-grained processing.}
    \label{fig:pipeline}
\end{figure}

Our system implements a multi-stage cascading RAG pipeline combining LLM-based conversational query rewriting, learned sparse retrieval, pointwise and LLM listwise reranking, and final response generation. Figure~\ref{fig:pipeline} presents the overall architecture, and each component is detailed below. The overall idea is to use a cascading ranking pipeline, with an effective method for retrieval, and then more expensive ranking approaches on the top retrieved passages for a more fine-grained final ranking.

\header{Query Rewriting.}
We first apply LLM-based conversational query rewriting to transform the context-dependent user utterance into a standalone query. Query rewriting enables the use of standard retrieval models by resolving anaphora and ellipsis while preserving the primary information need expressed in the dialogue. Below is the same example from the MTRAG paper~\cite{Rosenthal2026MTRAGEval}:
\vspace{0.5em}

{
\centering
\fbox{%
\begin{minipage}{0.95\linewidth}
\small
{User:} Who is the CEO of Apple Inc.?\\
{Agent:} The CEO of Apple Inc. is Tim Cook.\\
User: its address?\\
\\
\textbf{[Rewriting] What is the address of Apple Inc?}
\end{minipage}%
}
}

\header{Retrieval and Initial Reranking.}
The rewritten query is used for retrieval with the first-stage retrieval model, LION-SP~\cite{hansilion}, an LSR model trained on an LLM backbone. LION-SP, similar to SPLADE~\cite{splade++} combines neural semantic modeling with lexical sparsity, providing strong out-of-domain generalization while remaining compatible with efficient inverted-index search. This reduces the candidate set from the full collection to the top 1000 passages, which is then reranked with a pointwise reranking stage to obtain the top 20 passages.

\begin{table*}[t!]
\centering
\adjustbox{max width=\textwidth}{
\begin{tabular}{lccc}
\toprule
\multirow{2}{*}{\textbf{Task A - Retrieval}} 
& \multicolumn{3}{c}{\textbf{nDCG}} \\
\cmidrule(lr){2-4}
& @1 & @5* & @10 \\
\midrule
(best baseline) GPT-OSS-20b QR + ELSER & -- & 0.4795 & -- \\
\midrule
LSR w/ LION-SP-8B (retrieval)
& 0.4910 
& 0.4841 
& 0.5343 \\
\hspace{0.8em} + Qwen3-Reranker-8B (pointwise)
& 0.5120 
& {0.5477} 
& 0.5921 \\
\hspace{2.4em} + GPT-4.1 Listwise Reranking ($\dagger$) 
& \underline{0.5331} 
& \underline{0.5475} 
& \underline{0.5943} \\
\hspace{3.1em} \textit{\textbf{(Rank 2 out of 38)}} \\
\bottomrule
\end{tabular}
}
\caption{Retrieval performance of our submission on Task A of the MT-RAG SemEval 2026 MTRAG Task 8. We include ablations of the different steps of our pipeline and the best baseline from the organizer (GPT-OSS query rewriting with dense retrieval). ($\dagger$) denotes our submission \textbf{\textit{uva-1}}. (*) denotes the official metric used to compare participants.}
\label{tab:main_results}
\end{table*}

\header{Listwise Conversational Reranking.}
While rewriting improves retrieval compatibility, it compresses conversational context and may omit constraints or information introduced in earlier turns. To reintroduce this context, we perform a second reranking stage using LLM-based listwise ranking conditioned on the full dialogue history~\cite{sun2023chatgpt}. This stage jointly compares candidate passages and reorders the top 20 results into a final top 10, enabling finer-grained, context-aware evidence selection prior to generation. We only apply listwise reranking on a shorter retrieved passage list since usual context windows do not allow for including the full 100 or 1000 passages at once. By passing the top 20 we ensure that the model focuses on the most relevant passages, improving precision.

\header{Response Generation.}
Finally, the top 5 ranked passages are provided to an LLM for response generation in a zero-shot retrieval-augmented generation setting~\cite{gao2023retrieval}. Limiting generation to the highest-ranked evidence reduces noise while preserving sufficient contextual coverage. The model conditions on both the selected passages and the full conversation history to produce grounded, context-aware answers.

\section{Experimental setup}

For retrieval, we use an LSR model, Lion-SP-8B\footnote{\url{hzeng/Lion-SP-8B-llama3-marco-mntp}}~\cite{hansilion}. For initial pointwise reranking, we employ Qwen3-Reranker-8B\footnote{\url{Qwen/Qwen3-Reranker-8B}}~\cite{yang2025qwen3}, a large LLM-based reranker. The second-stage listwise reranking is performed using GPT-4.1~\cite{achiam2023gpt}, which is also used for final response generation. We relied on GPT-4.1, as the strongest non-reasoning model from OpenAI, achieving high performances on other conversational shared tasks~\cite{mohikat23}.

Query rewriting,  response generation and listwise reranking components are used in a zero-shot setting without task-specific fine-tuning. Similarly, retrieval models are only trained on MSMARCO, but not finetuned for the specific domains of the task. Inverted index search is implemented with the \texttt{Seismic} library~\cite{bruch2024efficient}, enabling efficient search over sparse representations. We apply representation pruning with query and document thresholds of (400, 600) to balance effectiveness and efficiency (although retrieval is not the efficiency bottleneck of the method).

\begin{table}[]
\centering
\adjustbox{max width=0.49\textwidth}{
\begin{tabular}{lcccc}
\toprule
\textbf{Task A -} 
& \multirow{2}{*}{{ClapNQ}}
& \multirow{2}{*}{{FiQA} }
& \multirow{2}{*}{{Govt} }
& {IBM} \\
\textbf{Retrieval} & & & & {Cloud} \\
\midrule
\textbf{uva-1} (ours)
& 0.645 
& 0.330 
& 0.587 
& 0.552 \\
nb. turns & (84) & (59) & (106) & (87) \\
\bottomrule
\end{tabular}
}
\caption{nDCG@5 retrieval performance of our submission on the four domains of MTRAG.}
\label{tab:ndcg5_per_collection}
\end{table}

For the retrieval track, systems are evaluated using nDCG~\cite{ndcg}, with nDCG@5 as the main metric. For response generation, the organizers reported three metrics: $\mathbf{RB_{alg}}$, the harmonic mean of BERT-Recall, BERT-K-Precision, and ROUGE-L~\cite{adlakha2024evaluating}, $\mathbf{RB_{llm}}$, an LLM-based evaluation metric~\cite{kuo2025rad}, and $\mathbf{RL_F}$ measuring faithfulness using the RAGAS framework~\cite{es-etal-2024-ragas}. They also computed the harmonic mean of these three together to rank participants' submissions ($\mathbf{H.Avg}$). The organizers also split turns into answerable, partially answerable and unanswerable, based on the passage assessment made on the collection.

\begin{table*}[t!]
\centering
\adjustbox{max width=\textwidth}{
\begin{tabular}{l|cccc|cccc}
\toprule
System & \multicolumn{3}{c}{IDK -Conditoned*} & & \multicolumn{3}{c}{Not Conditioned} \\
 & {RB\_agg} & {RB\_llm} & {RL\_F}& {\textbf{H. Avg*}}  & {RB\_agg} & {RB\_llm} & {RL\_F} & {\textbf{H. Avg}} \\
\midrule
\multicolumn{4}{l}{\textit{\textbf{Task B - Generation w/ Oracle Retrieval}}} & & \\[0.3em]
uva-oracle  
& {0.3683} 
& {0.6807} 
& 0.5981 
& {\textbf{0.5123}}  & {0.3590} & 0.8280 & 0.5899 & 0.5274 \\[0.5em]
\multicolumn{4}{l}{\textit{(Rank 23 out of 26)}} & & \\
\midrule
\multicolumn{4}{l}{\textit{\textbf{Task C - Generation w/ Predicted Retrieval}}} & \\[0.3em]
uva-rag 
& 0.3197 
& 0.6538 
& {0.6626}
& \textbf{0.4865}  & 0.3455 & {0.8332} & {0.8035} & {0.5619} \\[0.5em]
\multicolumn{4}{l}{\textit{(Rank 20 out of 29)}} & &\\
\bottomrule
\end{tabular}
}
\caption{Response generation performance of our submission on Task B and C of the MT-RAG SemEval 2026 MTRAG Task 8 (\textbf{\textit{uva-oracle}} and \textbf{\textit{uva-rag}}). (*) denotes the official
metric used to compare participants (IDK-Conditioned H.Avg).}
\label{tab:overall_results}
\end{table*}

Overall the dataset contains 332 turns evaluated for retrieval, resp. 84, 59, 106 and 87 turns on ClapNQ, FiQA, GovT and IBM-Cloud subsets.

\section{Results}

\header{Task A - Retrieval.} Table~\ref{tab:main_results} reports the retrieval performance of our submission. To better isolate the contribution of each component, we analyze the retrieval and reranking stages separately.

The main performance gains stem from the retrieval stage and the first (pointwise) reranking step. This implies that query rewriting with GPT-4.1 already provides a strong foundation. Our retrieval alone also achieves an nDCG@5 of 0.4841, which surpasses the official baseline based on GPT-OSS-20B query rewriting combined with ELSER, which reports an nDCG@5 of 0.4795. While this comparison suggests the strength of our rewriting and retrieval setup, the recall@1000 of the baseline system is not reported, which limits deeper analysis of candidate coverage.

Using LION-SP-8B retrieval alone yields a strong nDCG@1 of 0.4910. The subsequent pointwise reranking step improves early precision by +2.1 nDCG@1 points. Finally, the listwise LLM reranking stage provides an additional +2.1 point gain at rank 1. Notably, this second reranking stage primarily benefits top-ranked results (nDCG@1), while yielding only marginal improvements at nDCG@5 and nDCG@10.
Overall, our final nDCG@5 of 0.5475 places our system 2nd out of 38 submissions, with nDCG@5 serving as the official evaluation metric of the shared task.

\begin{figure}[t!]
    \centering
    \includegraphics[width=\linewidth]{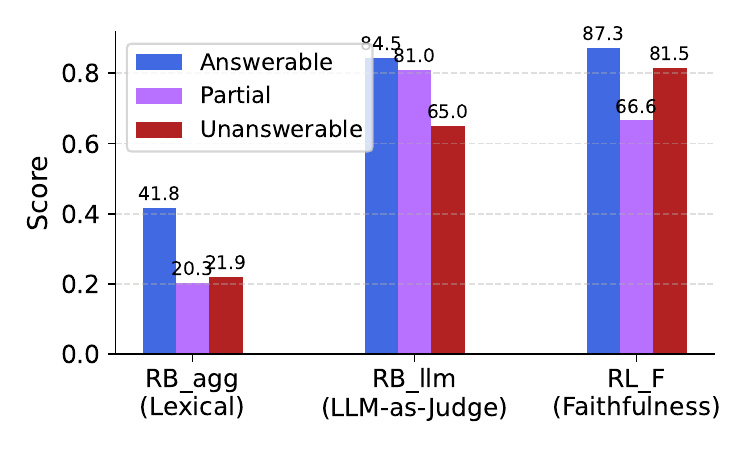}
    \caption{
    Response generation performance for different levels of answerability.}
    \label{fig:answerable}
\end{figure}

\header{Domain-Specific Results.} We then report in Table~\ref{tab:ndcg5_per_collection} the performance of our submission across domains. Consistent with observations from the original MT-RAG benchmark, \textit{FiQA} emerges as the most challenging collection among the four domains. Compared to the others, performance on FiQA is approximately half as high.
Despite this relative drop, the results remain competitive. On average, the first relevant passage for FiQA appears at rank 3, whereas for the other domains it appears around rank 2. This indicates that, although ranking quality is lower for FiQA, relevant evidence is still retrieved early in the ranked list.

\header{Task B and C - Response Generation.} Table~\ref{tab:overall_results} presents the performance of our RAG submissions using two answer quality metrics (RB\_agg and RB\_llm), a faithfulness metric (RL\_F), and their harmonic mean (H. Avg.). Results are presented separately for Task B and Task C. The official scores correspond to the evaluation conditioned on the IDK judge (left side of the table), which excludes underspecified turns.

Task C reflects a standard RAG setting: we use the retrieved passages from our Task A submission (uva-1) as context to generate answers within the conversational setting (uva-rag). In contrast, Task B serves as an oracle upper bound, where the generation model receives gold passages as context (uva-oracle).
As expected, the oracle configuration achieves higher answer quality scores, indicating that access to gold evidence improves content relevance and completeness. However, our RAG pipeline achieves higher faithfulness (RL\_F). We attribute this to the alignment between the reranking step and the final answer generation model, as both rely on the same LLM. In the oracle setting, although gold passages are provided, the generation model may selectively omit parts of the evidence or filter information it deems less relevant, which can reduce faithfulness.

\begin{figure}[t!]
    \centering
    \includegraphics[width=\linewidth]{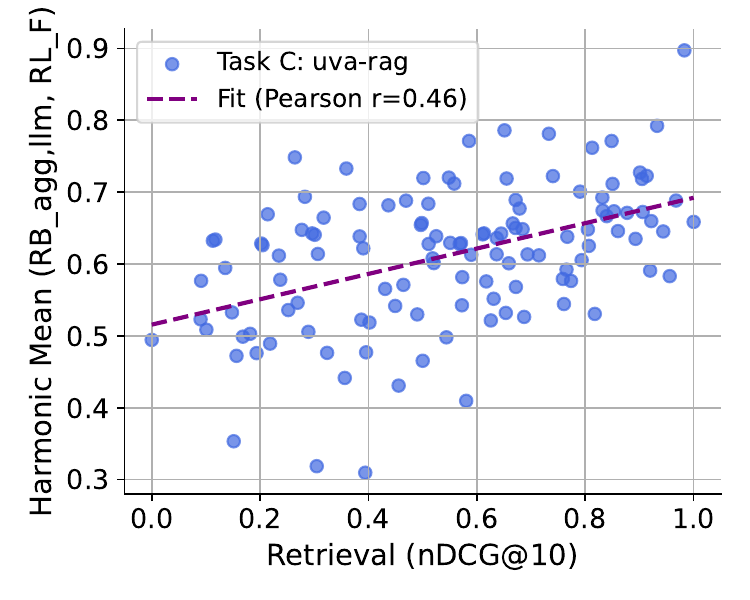}
    \caption{
    Retrieval and response generation correlation on partial and answerable turns. Pearson values for RB\_agg, RB\_llm and RL\_F are 0.48, 0.41 and 0.08.}
    \label{fig:scatter}
\end{figure}

\header{Partial and Non-Answerable Turns.} The organizers annotated a subset of turns as either non-answerable or partially answerable, requiring systems to explicitly identify underspecified cases. This results in two versions of each metric: the conditional variant (\_idk\_underspecified) and the original, unconditioned scores (the right section of Table~\ref{tab:overall_results} reports the latter).
When not conditioning on the IDK judgment, performance increases substantially. In particular, the harmonic mean reaches 0.5619, representing an 8-point improvement over the conditioned evaluation. Under this setup, our RAG pipeline (Task C) outperforms the oracle configuration (Task B), which further emphasizes the strength of our retrieval component. Notably, we observe a faithfulness score of 0.8035, indicating strong alignment between generated answers and retrieved evidence.

Figure~\ref{fig:answerable} further breaks down performance across answerable, partially answerable, and non-answerable subsets (without IDK conditioning). While answer quality metrics (RB\_agg and RB\_llm) decrease for partially or non-answerable turns, the model remains relatively faithful even for unanswerable questions. This suggests that, although content completeness is affected, the system generally avoids hallucinating unsupported information. 
We did not explore in more detail why our submission achieved such a high faithfulness, this could be attributed to a conservative prompt for the RAG component. For more details, prompts of our submission are included in Appendix~\ref{sec:appendix}.

\begin{figure}[t!]
    \centering
    \includegraphics[width=\linewidth]{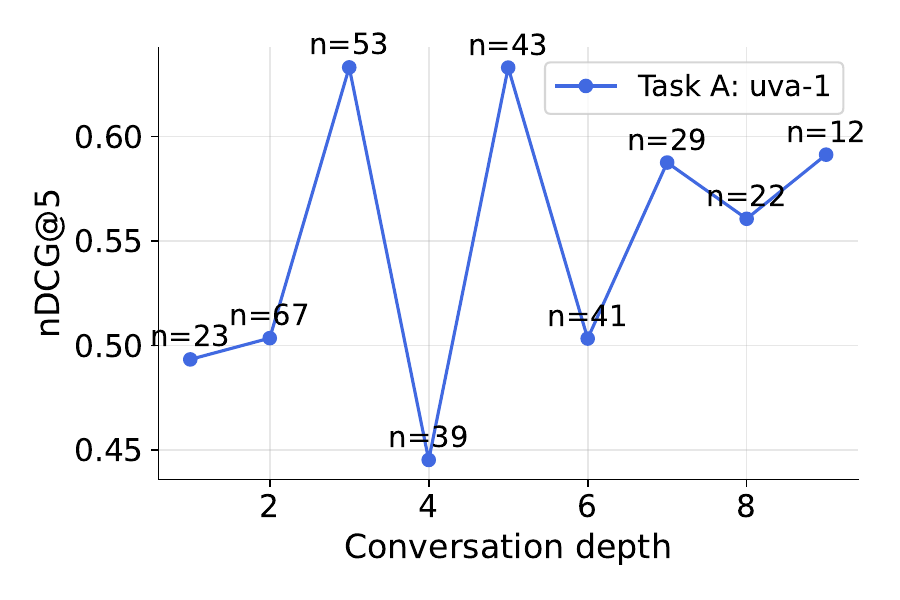}
    \caption{
    Retrieval Performance at varying depths.}
    \label{fig:taska-depth}
\end{figure}

\begin{figure}[t!]
    \centering
    \includegraphics[width=\linewidth]{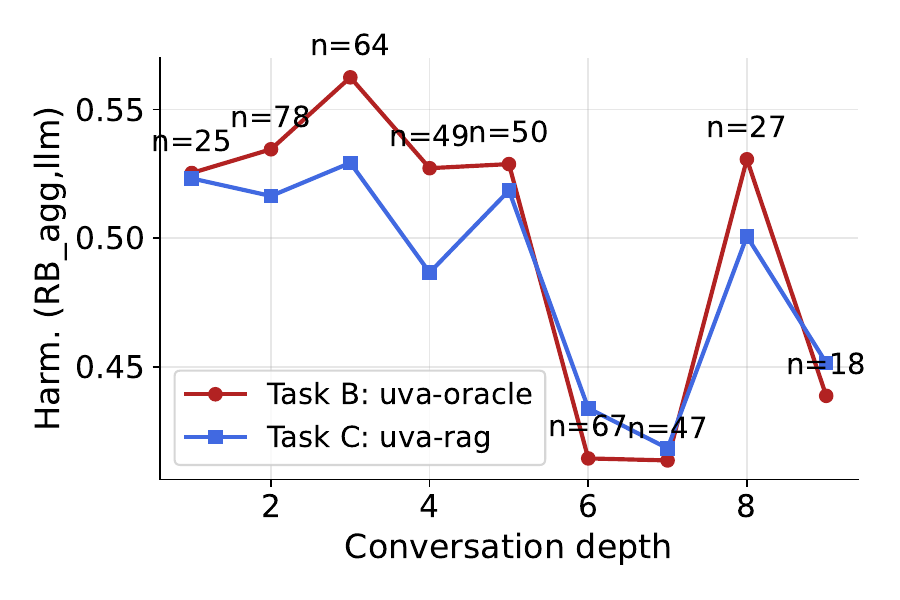}
    \caption{
    Response Generation at varying depths.}
    \label{fig:taskbc-depth}
\end{figure}

\header{Correlation Retrieval and Response Generation.} Figure~\ref{fig:scatter} illustrates the relationship between retrieval effectiveness and response generation quality. For this analysis, we consider only answerable and partially answerable turns, and report results without conditioning on the IDK judge. We observe a positive correlation, with a Pearson correlation between retrieval and generation metrics of 0.46, indicating that strong retrieval is generally associated with improved response quality.
When computing the Pearson correlation between nDCG@10 and each generation metric individually, we find that the reference-based metrics (RB\_agg and RB\_llm) exhibit positive correlation with retrieval effectiveness. In contrast, faithfulness (RL\_F) shows no correlation with nDCG@5. This is expected, as faithfulness measures alignment with the provided evidence rather than agreement with a gold reference answer, and is therefore less directly dependent on retrieval ranking quality.

\header{Performance by Conversation Depth.} Finally, we analyze performance as a function of conversational depth across all tasks. For retrieval (Task A), Figure~\ref{fig:taska-depth} shows no clear degradation as depth increases, suggesting that the query rewriting and retrieval components handle longer conversational histories consistently.
In contrast, for response generation (Tasks B and C), Figure~\ref{fig:taskbc-depth} reveals a performance decline with increasing depth.
For generations, we report the harmonic mean of the reference-based metrics only (excluding faithfulness), as we assume faithfulness to be independent of conversational depth.

\section{Conclusion}

We present in this report an effective ranking pipeline for conversational search that combines query rewriting and full conversation context modeling with a learned sparse retrieval backbone. 
Our analysis further shows a positive correlation between retrieval effectiveness and response quality, confirming the importance of ranking for downstream generation. At the same time, faithfulness appears less directly tied to retrieval metrics, highlighting its distinct role in evaluating grounded responses. We also observe that our submitted system remained largely faithful even in challenging cases. Although performance on reference-based metrics decreases for partially or non-answerable queries, the system generally avoids introducing unsupported information, even if it does not always explicitly acknowledge missing knowledge.

\section{Acknowledgments}

This research was partly supported by the Swiss National Science Foundation (SNSF), under the project PACINO (Personality And Conversational INformatiOn Access), grant number 215742.

\bibliography{custom}

\appendix

\newpage

\section{Prompts}
\label{sec:appendix}

We provide in this section the detailed prompts we used for our submission, including LLM query rewriting, listwise LLM reranking and response generation. All prompts were used on GPT-4.1.

\begin{table}[h]
    \begin{lstlisting}
You are a query rewriting model for conversational information retrieval. 
Rewrite ONLY the final user turn into a single, standalone search query. Use earlier turns ONLY to resolve references, ellipsis, and ambiguity.

Guidelines:
- Maximize lexical clarity and keyword recall.
- Prefer explicit noun phrases over pronouns.
- Preserve important domain terms from all turns.
- Do NOT answer the question.
- Do NOT add new facts or assumptions.
- Do NOT explain or comment.
- Output exactly one query.

Output ONLY the rewritten query text.
Conversation: {conversation_text}
Rewrite the last user message into a standalone query.
Last user message: {last_user_utterance}
    \end{lstlisting}

\caption{Query Rewriting prompt to obtain a single standalone query from a full context conversation.}
\label{tab:prompt_rw}
\end{table}

\begin{table}[t]
    \begin{lstlisting}
You are an expert relevance judge for conversational search.
Conversation: {conversation_text}

Task:
Rank the following passages by how useful they are for answering the user's current information need. The full conversation provides context, but prioritize the latest user turn.

Guidelines:
- Prefer passages that directly answer the current user need.
- Use earlier turns only to resolve references or constraints.
- Penalize passages relevant only to earlier turns.
- Do NOT generate an answer.

Passages: {top20_retrieved_text}
Return exactly 10 passage labels ranked from best to worst, using only labels (D1..D20).
Example: D1 > D20 > D7 > ... > D14
    \end{lstlisting}
      \caption{LLM Listwise reranking prompt.}
  \label{tab:prompt_rerank}
\end{table}

\begin{table}[t]
    \begin{lstlisting}
You are a conversational question answering assistant. Use the conversation history to understand context and intent. Answer the LAST user question using the information in the documents. Answer in maximum 256 tokens.

Conversation: {conversation_text}
Retrieved Documents: {top5_retrieved_text}
    \end{lstlisting}
      \caption{Response Generation with RAG prompt.}
  \label{tab:prompt_rag}
\end{table}

\end{document}